\documentclass{article}

     \PassOptionsToPackage{numbers, compress}{natbib}
     \usepackage[preprint]{neurips_2020}

\usepackage[utf8]{inputenc} % allow utf-8 input
\usepackage[T1]{fontenc}    % use 8-bit T1 fonts
\usepackage{hyperref}       % hyperlinks
\usepackage{url}            % simple URL typesetting
\usepackage{booktabs}       % professional-quality tables
\usepackage{amsfonts}       % blackboard math symbols
\usepackage{nicefrac}       % compact symbols for 1/2, etc.
\usepackage{microtype}      % microtypography

\usepackage{graphicx}
\usepackage{subfigure}
\usepackage{amsmath}
\usepackage{amsfonts}
\usepackage{wrapfig}

\title{Implications of Human Irrationality for Reinforcement Learning}
\author{%
  Haiyang Chen \qquad Hyung Jin Chang \qquad Andrew Howes\\
  School of Computer Science\\
  University of Birmingham\\
  Birmingham, United Kingdom\\
  \texttt{\{h.chen.7, h.j.chang, howesa\}@bham.ac.uk} \\
}

\begin{document}

\maketitle

\begin{abstract}
 Recent work in the behavioural sciences has begun to overturn the long-held belief that human decision making is irrational, suboptimal and subject to biases. This turn to the rational suggests that human decision making may be a better source of ideas for constraining how machine learning problems are defined than would otherwise be the case. One promising idea concerns human decision making that is dependent on apparently irrelevant aspects of the choice context. Previous work has shown that by taking into account choice context and making relational observations, people can maximize expected value. Other work has shown that Partially observable Markov decision processes (POMDPs) are a useful way to formulate human-like decision problems. Here, we propose a novel POMDP model for contextual choice tasks and show that, despite the apparent irrationalities, a reinforcement learner can take advantage of the way that humans make decisions. We suggest that human irrationalities may offer a productive source of inspiration for improving the design of AI architectures and machine learning methods.
\end{abstract}

\section{Introduction}
\label{introduction}

%The human mind has long inspired ideas as to how to build intelligent machines in the Artificial Intelligence (AI) field by re-implementing the transfer of the insights gained from cognitive science, psychology, and neuroscience \cite{hassabis2017neuroscience,lake2017building}. 
Humans have long been a source of inspiration for how to build intelligent machines \cite{hassabis2017neuroscience,lake2017building}.
%In return, the development of AI has the potential to offer deeper understandings of the human mind by building computational models \cite{lewis2014computational,lake2017building,gershman2015computational}. 
There is now a series of successful examples of where knowledge about the brain and mind has been used to develop new types of Machine Learning (ML), including artificial neural networks \cite{mcculloch1943logical}, convolutional neural networks inspired in part by the hierarchical organization of vision \cite{lecun2015deep}, and Reinforcement Learning (RL) which was inspired by decision making and learning under uncertainty in humans and other animals \cite{littman2015reinforcement}. 
A more recent example is provided by the promise of the utility of uncertainty, which has demonstrated that incorporating human-like uncertainty about object classifications can help obtain more robust and better performing machine classification \cite{peterson2019human}. Many recent advances have come from modeling uncertainty in ML. For example, capturing uncertainty can improve model performance in regression and classification tasks \cite{kendall2018multi, kendall2017uncertainties}, estimating uncertainty can improve deep learning algorithms \cite{gal2016dropout, maddox2019simple, osawa2019practical}, and representing the uncertainty of an agent's policy can aid more efficient exploration in RL \cite{fortunato2017noisy,o2018uncertainty, janz2019successor}. Another example of the influence of the human sciences on ML is how selective attention in human perception and neural information processing, has motivated  rapid  progress in object recognition \cite{ba2015multiple}, visual object tracking \cite{Choi2016CVPR, Choi2017CVPR, Choi2018CVPR}, human action recognition \cite{Lee2015TIP}, image caption generation \cite{xu2015show}, and machine translation \cite{bahdanau2015neural,vaswani2017attention}. In sum, progress on multiple fronts suggests that human cognition offers a productive source of inspiration for improving ML.

% Contextual sequence processing such as language understanding in the human brain . (Context and Compositionality in Biological and Artificial Neural Systems)

% draw inspiration from concepts and findings in human cognition
% took root in ideas from the brain and mind
% using insight from human cognition

% Two part: 
% 1. help to improve ML. Neural-inspiration from cognitive science into ML
% 2. help to understand CogSci. explore or understand Cognitive science using ML as Scientific model.

However, perhaps not every aspect of human cognition should be emulated. One such aspect might be human choice behavior which has generated a long list of cognitive biases \cite{tversky1974judgment}, leading to systematic and predictable errors. For example, humans supposedly rely too heavily on the first piece of evidence gathered (the anchoring bias), they heavily weight decisions towards more recent information (the availability bias), and they overweight the value of outcomes that are expected to definitely occur (the certainty bias). At the last count, in a decision-making textbook, \cite{baron2008thinking} lists 53 irrationalities. Together, these biases have been taken to suggest that people are not good expected value maximizers and that they are subject to irrationalities of choice that are counter to their own self-interest. 

A highly critical appraisal of the ``heuristics-and-biases'' tradition has recently been provided by \cite{gigerenzer2018bias}. Indeed, recent research has begun to show that people may be more rational than supposed \cite{lewis2014computational,chen2015optimal,lieder2019resource,todd2012ecological,frazier2008sequential}. 
%We believe that there is much for ML to learn from why apparent biases arise in humans. 
%Even though  biases may appear to lead to systematic deviation from optimal decision making, they may also be more economical and effective \cite{tversky1974judgment}. 
%A better understanding of biases (also known as irrationalities) could improve machine decision making under various conditions, especially under uncertainty. 
One type of bias in humans is known as a context-of-choice bias in which irrelevant options influence decisions. For example, if a person chooses an apple over a cake on the grounds of health, but then chooses the same cake when the choice is between an apple, the cake and another cake with extra sugar, then the clearly inferior (on health grounds) ``cake with extra sugar'' has influenced the choice between two superior alternatives. 
This is an example of where human choice between two options changes when a third (dominated) alternative is introduced and there is significant empirical evidence demonstrating this phenomenon \cite{wedell1991distinguishing}. As the dominated choice is irrelevant to the choice between the other two options, it should have no effect on their valuation, nor on the choice. This effect has been taken by some as evidence that human cognition is irrational since it appears to violate the normative principles of independence.  \cite{usher2001time,roe2001multialternative,tversky1993context}. 
However, \cite{howes2016contextual} has shown that these apparent irrationalities can emerge from computationally rational mechanisms. 

%These contextual decision making effects have widely been taken as evidence of irrationality \cite{tversky1993context}. 

In the current paper, we propose an agent inspired by \cite{howes2016contextual} demonstration that apparent irrationalities of choice can emerge from rational processing. Our approach  is an example of a broader class of analysis known as Computational Rationality \cite{lewis2014computational,gershman2015computational,lieder2019resource}.
It extends  \cite{howes2016contextual} by modeling contextual choice tasks as sequential decision problems and formulating them as Partially Observable Markov Decision Processes (POMDPs). 
Previous work by \cite{dayan2008decision,rao2010decision,frazier2008sequential, lieder2017automatic,oulasvirta2018computational} and others has established the value of POMDPs and related formalisms for modeling humans.
In our work, a reinforcement learning agent, designed to solve a POMDP, acquires a sequential decision policy that chooses what information to gather about which options, calculates option values, and makes comparisons between options as the unfolding task demands. The agent is trained and tested on sampled choices between three gambles, each expressed as a probability and a value.
The agent learns the relative value of (1) noisy calculation of option values (e.g. by multiplication of a probability by a value), (2) noisy comparisons (e.g. comparing two probabilities to see which option is riskier), and (3) acting (making a choice). The agent is not pre-programmed to gather all information but learns to gather only that information that helps it maximize utility. We contrast this agent to other simpler agents and show that the human-inspired agent performs better (achieves higher cumulative reward) than an agent that makes independent assessments of each option value, replicating the results of \cite{howes2016contextual} but in the POMDP setting.

Our analysis of the new agent's learned policy shows that it \emph{learns} to use contextual information to help infer which options approximately maximize expected value while taking into account computational cost and cognitive limits. The agent's performance shows that making use of contextual information helps it make more accurate and efficient decisions under uncertainty while also giving rise to apparently irrational and human-like decision making \cite{howes2016contextual}. The work demonstrates that, under choice uncertainty, there is economic value to ML agents of policies that make use of choice context and relational judgments. 
%This observation suggests ways in which ML research with directions to make decisions more human-like and to be better aligned with human preference.

%In addition, our human-inspired model predicts when, and explains why, people stop evidence accumulation and make a decision. It might suggest the following specially designed experiment with humans, which would help understand the cognitive process mechanism underlying the irrationality and give rise to new findings of human cognition.

% \textbf{Contributions:} 

This paper's contributions are as follows:

\begin{itemize}
\item It replicates the previous findings of \cite{howes2016contextual} with a new problem formulation based on POMDPs;  it shows that preference reversals emerge from learning in a POMDP setting as well as in the exhaustive search setting reported by \cite{howes2016contextual}.
\item It provides further evidence that cognitive biases can inform ML. Model simulations demonstrate that it is more profitable for an RL agent to take into account context and relational judgments when choosing between uncertain options than to make independent evaluations of each option, suggesting that in some circumstances \emph{RL agents should be designed with the capacity to compare options}.
\item It extends the analysis of \cite{howes2016contextual} to account for the impact of information gathering costs on contextual choice. 
\item It makes novel predictions concerning optimal sequential information gathering in contextual choice tasks. In particular, it shows how the ratio of option comparisons and expected value calculations is influenced by the level of uncertainty in the observation functions.
%\item It presents, to our knowledge, the first supporting evidence for the utility of the computational rationality framework based on POMDPs for modeling contextual effect in choice tasks. It extends previous work showing how human contextual biases can be approximately rational  \cite{howes2016contextual} to problem domains formulated as POMDPs. 

%\item It  shows  how  human  cognitive  biases  inform   Model simulations demonstrate that it is more profitable for an RL agent to take into account context and relational judgments when choosing between uncertain options than to make independent evaluations of each option. 
%AH Feb 6th DELETE: It reveals the economic value of using contextual information in helping humans or agents make decisions accurately and efficiently in the situation of uncertainty.
%\item It introduces a new approach to modeling human contextual choice behavior and makes quantitative predictions that correspond well to evidence about human choice. 
%AH Feb 6th DELETE: The resulting model provides a promising way to explain human decision making by integrating the insights from cognitive ingredients and ML methods.
%\item It demonstrates that apparent cognitive biases can emerge %from approximately computationally rational processes, formulated as POMDPs. 
%It thereby adds to the growing literature supporting computational rationality as a useful paradigm for cognitive science and AI. 
%\item perhaps add something about human-centred AI.
\end{itemize}

The paper is organised as follows, we first describe experiments revealing contextual choice effects in human decision making and review theoretical accounts of human choice from the cognitive science and neuroscience perspectives. We then define a contextual choice problem as a POMDP that includes ``comparison'' observations and describe how to solve this POMDP with a reinforcement learning agent.
%that is based on cognitive and neural principles. 
We test the agent on gamble tasks for which humans are known to be influenced by choice context, and we demonstrate the correspondence between the approximately reward maximizing RL behavior and human behavior - replicating \cite{howes2016contextual}. Lastly, we analyse the agent's sequential behaviour and reveal the effect of observation noise on the frequency with which it uses option comparisons.
%explain several core aspects of mechanisms leading to context effects and discuss the insight on understanding human cognition as well as improving ML.

\section{Background}
\subsection{The Effect of Choice Context on Humans}

%Before formally describing the model, we introduce three particular types of context effect that have had a significant role in shaping cognitive and neural theories  of human decision making \cite{busemeyer2019cognitive,wollschlaeger2019similarity}.

%\subsection{Context Effects as Examples of Human Irrationality}

\begin{figure}[b]
\centering
\includegraphics[width=0.99\columnwidth]{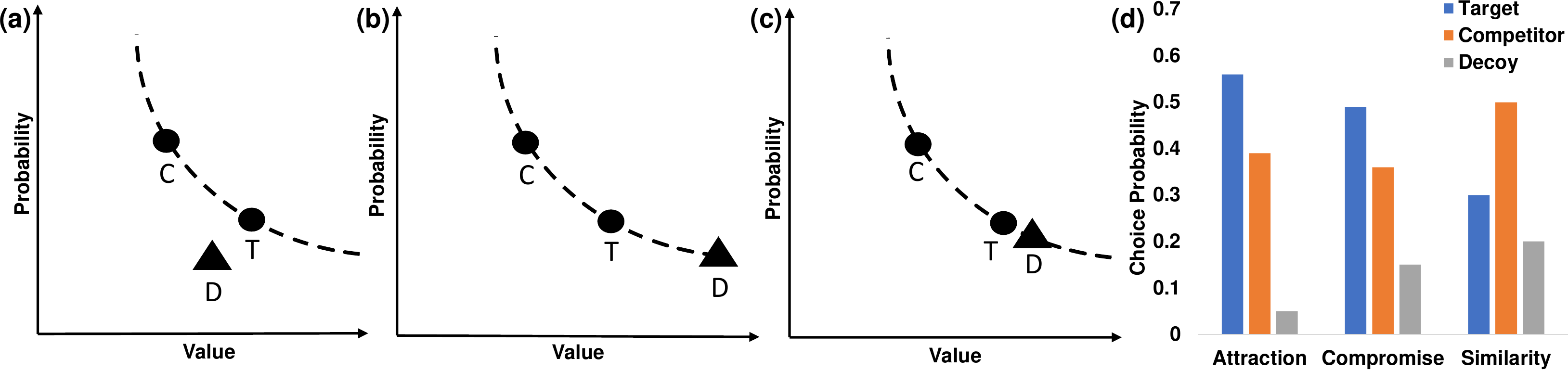} 
\vspace{-2mm}
\caption{ (a)(b)(c) An illustration of the options in three types of contextual choice task -- called the attraction (a), compromise (b) and similarity (c) tasks. The Target \(T\) and Competitor \(C\) are two options and have equal expected value (the dotted line). Option \(D\) is a decoy designed for comparison with the Target \(T\). In the attraction task (a), \(T\) dominates \(D\). In the compromise task (b), \(T\) is a compromise between \(D\) and \(T\). In the similarity task (c), \(D\) has similar expected value to \(T\). (d) Proportion of choices of each of the three options (Target, Competitor and Decoy) in each of the the three contextual choice tasks (Attraction, Compromise and Similarity). The Target is preferred in the Attraction and Compromise tasks and the Competitor is preferred in the Similarity task. Data are reproduced from \cite{trueblood2012multialternative}. } 
\label{position}
\end{figure}

As we have said, the human behaviours that influence this paper are those exhibited in decision-making tasks in which people appear biased by seemingly irrelevant context. Here we look in more detail at these tasks and their effects. Contextual decision experiments have revealed many robust empirical effects and contributed to shaping cognitive  theories of human decision making \cite{usher2001time,roe2001multialternative,howes2016contextual,busemeyer2019cognitive,noguchi2018multialternative,ronayne2017multi,wollschlaeger2019similarity}.
Three of the most well known contextual decision task are the attraction, compromise and similarity tasks. These are illustrated in Figure \ref{position}a, b, c. For the attraction type task, there are two best options (the Target and the Competitor) with the very similar expected value. Each option  is best on one dimension but not the other. One of these two options (the Target) dominates a third option, called the decoy, on both dimensions.
It is difficult to choose between the two best options since each option dominates the other on one of the attributes. 
Experiments studying these three tasks have been reported by many authors. Consider the results of an experiment in which participants were asked to make decisions about criminal suspects \cite{trueblood2012multialternative}. Participants were presented with a sequence of tasks each consisting of three suspects and were asked to decide which suspect was most likely to have committed a crime. 
There were two types of evidence, of varying strength, about each of the three suspects, such that the suspects had likelihoods of criminality in patterns identical to the three patterns presented in Figure \ref{position}d. These three patterns were used as the materials in the three conditions of the experiment.

In the attraction condition of the experiment, there were two equally likely criminal suspects and a decoy suspect who was less likely than the other two (Figure \ref{position}a). The experimental results showed that the Target suspect who dominates the decoy was chosen more frequently  than the Competitor suspect. 
%This is known as the attraction effect (Figure \ref{humanPreference}).
In the compromise condition of the experiment (Figure \ref{position}b) the findings showed that the suspect who is in-between the Competitor and decoy is chosen more frequently than the Competitor. 
%This is known as the compromise effect (Figure \ref{humanPreference}).
In the similarity condition (Figure \ref{position}c), the results showed that the suspect who is very similar to the decoy is chosen less frequently than the Competitor. %This is known as the similarity effect (Figure \ref{humanPreference}).

%In the attraction task, the decoy is chosen infrequently (less than 5 percent of the time in most experiments). In the similarity and compromise tasks, the decoy is chosen much more frequently. A human-like decision model must produce the same pattern of data in Figure \ref{humanPreference}.

Human behaviour on these tasks has been seen as biased because the sensitivity to irrelevant context (the decoy option) appears to have consequences for the choice between the other two options(\cite{tversky1993context}, p. 1188). 
%Tversky and Simonson (\cite{tversky1993context}, p. 1188) state, “The analysis of context effects, in perception as well as in choice, provides numerous examples in which people err by complicating rather than by simplifying the task; they often perform unnecessary computations and attend to irrelevant aspects of the situation under study.” 
%More recently, Usher, Elhalal, and McClelland (\cite{usher2008neurodynamics}, p. 297) state, “... contextual reversal effects ... demonstrate a limitation of rationality in choice preference.” Simonson (\cite{simonson2015mission}, p. 1) states, “the belief in irrationality is now widely accepted among the general public. 
The most commonly used operationalization of irrationality among decision researchers has been based on violations of value maximization. Preferring a dominated option or expressing different preferences depending on the framing of options demonstrates irrational decisions. The significance of any irrationality, if that is what they are, cannot be understated  given the potential for catastrophic real world consequences.
%in contexts such as criminal justice.  
However, the conclusion that choice under uncertainty provides evidence of irrationality may be incorrect \cite{howes2016contextual,tsetsos2016economic}. Substantive analysis of the value of comparing options has shown that they are in fact informative and are required, under conditions of uncertainty, for reward maximization \cite{howes2016contextual}. The substantive structure of these analyses has informed the design of the agent that we present below. The key cognitive strategy that is borrowed from human behaviour is the use of option comparison to inform decision making under uncertainty. Comparison was extensively explored by Stewart \cite{stewart2006decision, vlaev2011does} who has documented extensive of its use in a range of human decision making tasks. For example, there is eye tracking evidence \cite{noguchi2014attraction} that people tend to make more eye movements that switch between options than eye movements that gather all of the evidence about a single option; evidence which is consistent with the use of comparisons.

\subsection{Human Decision Making as a POMDP}

POMDPs provide a mathematical framework for sequential decision processes \cite{kaelbling1998planning}.
POMDPs have previously been used for modelling and explaining various aspects of human decision making \cite{daw2006representation, dayan2008decision, rao2010decision}. An early contribution was \cite{daw2006representation}'s model of the dopamine system which incorporated semi-Markov dynamics and partial observability. \cite{rao2010decision} proposed a model of neural information processing based on POMDPs and tested this model on perceptual tasks such as the random dot motion task. Further work in perceptual decision making, has used the POMDP framing to explore  model confidence \cite{khalvati2015bayesian} and understand the role of priors \cite{huang2012prior}. POMDPs have even been used to model social decision making \cite{khalvati2016probabilistic}. More recently, meta-level Markov decision processes (meta-MDP), a closely related framework, have been used for modelling higher level decision making  \cite{griffiths2019doing}. The Meta-MDP model is similar to the belief-MDP version of the POMDP, but replacing physical actions with cognitive operations. Meta-MDPs have been used to model strategy selection and heuristics in decision making \cite{lieder2017automatic} and attention allocation in perception \cite{callaway2020fixation}.

Contextual preference reversals have influenced a number of models of human decision making \cite{usher2001time, roe2001multialternative, busemeyer2019cognitive, frazier2008sequential,trueblood2014multiattribute,ronayne2017multi,noguchi2018multialternative,busemeyer2019cognitive}. Many of these models have focused on neurally plausible sequential processing, capturing the fact that decision  making usually requires accumulation of evidence and integration of information across time  \cite{tsetsos2016economic}. Other models  have focused on the way that people solve this problem by sampling comparisons between option attributes and thereby impose a rank order on options \cite{noguchi2018multialternative}. However, none to our knowledge, have shown that preference reversals are an emergent consequence of an RL solution to a POMDP.

%In summary, the POMDP framework has show the ability and advantage to model central aspects of human decision making including: evidence accumulation via information-gathering actions selection, threshold to reach a decision, decision time, noisy observation of the environment, costs and rewards of actions, etc. 

% Advantage of POMDP to formulate human decision making (choice): 1. sequential nature. 2. evidence accumulation via action selection. information gathering policy.  capture the dynamic of decision making tasks. Time pressure experiment. evolution driven. 3. no need to assume stop rule or threshold. 4. incorporate various uncertainty. 5. cost and reward of the actions. this is also no mentioned in Howes, et al 2016. But it is corespond to Dopamine system and important to understand human decision making and the underlying neural processes.   
% AH: I don't think that we should get into Dopamine.
% HY: Yes, I agree. Dopamine is too far from the topic. 

%\section{A Human-like Decision Making Agent}
\section{Contextual Choice as a POMDP}

We view contextual choice tasks as sequential decision making problems and formulate them as POMDPs that include, in the action space, comparison actions to assess choice option values. Given this formulation, we use a deep reinforcement learning model to discover an approximately optimal choice policy and demonstrate its capacity to simultaneously maximize reward and model humans. A crucial property of the model is that gathering information is costly, so that more information costs more but also increases the probability of a better, more rewarding, choice.

We start with a standard definition of a POMDP as a tuple \((\mathcal{S}, \mathcal{A}, \mathcal{O}, \mathcal{T}, \mathcal{Z}, \mathcal{R}, \gamma)\), where \(\mathcal{S}\) is the state space, \(\mathcal{A}\) is the action space, and \(\mathcal{O}\) is the observation space. At each time step \(t\) the agent is in the latent state \(s_t \in \mathcal{S} \), which is not directly observable to the agent. When the agent executes an action \(a_t \in \mathcal{A} \), the state of the process changes stochastically according to the transition distribution, \(s_t \sim T(s_{t+1}|s_t, a_t)\). Then, to gather information about the state, the agent makes a partial observation \(o_{t+1} \in \mathcal{O}\) according to the distribution \(o_{t+1} \sim \mathcal{Z}(o_{t+1}|s_{t+1}, a_t)\). The agent received a reward \(r_{t+1}\in \mathcal{R}\) according to the distribution \(r_{t+1} \sim \mathcal{R}(o_{t+1}|s_{t+1}, a_t)\) after performing an action \(a_t\) in a particular state \(s_{t+1}\). The agent must rely on its observations to inform action selection since the hidden states are not directly observable. In each time step \(t\), the agent acts according to its policy \(\pi(a_t|h_t)\) which returns the probability of executing action \(a_t\), and where \(h_t = (o_0,a_0,o_1,a_1,\cdots o_{t-1}, a_{t-1})\) are the histories of observations-actions pairs.
The goal of the agent is to learn an optimal policy \(\pi^*\) that maximizes the expected cumulative rewards,
\(
\pi^* = \underset{\pi}{argmax} \mathbb{E}\left [ \sum_{t=1}^{t=T}\gamma^{t-1}r_t \right ]
\)
, where \(0<\gamma<1\) is the discount factor.

Each choice task had 3 options (X, Y, Z) which were represented with two attributes: a randomly sampled probability \(p\) and a randomly sampled value \(v\). We assumed that probabilities \(p\) were sampled from a \(\beta-\)distribution and values \(v\) were sampled from a \(t-\)distribution. These distributions represented the ecological distributions experienced by participants in the human behaviour experiments reported by \cite{wedell1991distinguishing}. We view contextual choice tasks as sequential decision making problems and formulate them as POMDPs as follows.

%The state space \(\mathcal{S}\) for each task was generated from a sampled choice task. The agent's state  \(s_t\) describes a set of utility of one item. There are two utilities, representing  expected utility and comparison utility of the options. The state updated according to the action.

%The latent state space \(\mathcal{S}\) consisted of the set of all possible comparisons (of \(p\)s and \(v\)s) and calculations (p x v) and an index, set by the action, indicating which comparison or calculation was to be observed next. 

The state space \(\mathcal{S}\) for each task was generated from a sampled choice task.  More formally a state was \(\{(p_X, v_X)\), \((p_Y, v_Y)\), \((p_Z, v_Z)\}\), where probabilities \(p\) were sampled from a \(\beta-\)distribution and values \(v\) were sampled from a \(t-\)distribution. The agent selected actions from a set \(\mathcal{A}\) which included 6 comparison actions (e.g. compute the comparison relation for \(p_X\) and \(p_Y\)), 3 calculation actions (e.g. compute the expected value for X given \(p_X\) and \(v_X\)) and the 3 choice actions (choose X, choose Y, choose Z). The reward for comparison and calculation actions was negative \(c\). The reward for a choice action was 10 if the agent chose the option with maximum expected value, otherwise, it was -10. 
%The cost (reward) of taking an action was \(c\), representing a fixed time cost of action. The task was terminated when the agent made a choice. The reward was 10 if the agent chose the option with maximum expected value, otherwise, it was -10. 
There was therefore a trade-off between the cost of information gathering and choice accuracy. More information cost more but was more likely to lead to a better response and therefore a higher reward.
As a consequence of the selected action, the subsequent observation \(o_{t+1} \in \mathcal{O}\) was of computing the most recent comparison or calculation with noise. 
Following \cite{howes2016contextual} each observation of a comparison had 4 possible outcomes, which indicated that the relation was unknown, greater, equal and less. 
%These are:
%\begin{equation} \label{eq2}
%f(m_i, m_j)=\begin{cases}none, & unknown\\ >, & m_i > m_j + \tau_m\\\equiv, & \left | m_i - m_j \right |\leq \tau_m\\ <, & m_i < m_j - \tau_m\\\end{cases}
%\end{equation}
%\[\]
The function \(f\) represents this pairwise order relation between the two values or two probabilities of two gambles.
%, the magnitude  \(m \in \{v, p\}\) is the magnitude of value or probability. The relation is defined as equal if their magnitudes \(m\) are within their corresponding tolerance \(\tau_m\). 
The probability of comparison error \(P(error_f)\) was the probability that the relations were sampled uniformly random from the comparison set \( O= \{>, \equiv, < \} \).
%\[\large O =\left\{f(p_A, p_B), f(p_A, p_D), f(p_B, p_D), f(v_A, v_B), f(v_A, v_D), f(v_B, v_D)\right\}\]
The observation of a calculation was computed using:
\begin{equation} \label{eq4}
 E_i = p_i^\alpha \times v_i + \varepsilon \qquad \varepsilon \sim N(0, \sigma_{calc}^2)
\end{equation}
where the probability \(p\) was weighted by an exponential parameter \(\alpha\). 
The purpose of using parameter \(\alpha\) is to model \textbf{subjective probability} following \cite{savage1972foundations}. The exponential is used to model \textbf{subjective probability} is extensively in econometrics because it is mathematically well behaved.

The evidence state is the history of the partial and noisy observation of the latent state. The history of observation set \(\mathcal{O}_h\) is the noisy encoding of the partial orderings of probabilities and values:
\begin{equation} \label{eq3}
\begin{split}
 \mathcal{O}_h =\{f(p_X, p_Y), f(p_X, p_Z), f(p_Y, p_Z), f(v_X, v_Y), f(v_X, v_Z), f(v_Y, v_Z), E_X, E_Y, E_Z \}
\end{split}
\end{equation}

% The observation \(o_t\) is the same as the latent state \(s_t\), when there is no noisy observation: the tolerance \(\tau_m = 0\),  the exponential parameter \(\alpha = 1\), the probability of comparison error \(P(error_f) = 0 \) and the calculation noise \(\sigma_{calc}=0\).

It is intractable to compute a policy to solve the defined POMDP, but it is possible to approximate the optimum through learning   \cite{cushman2015habitual,wang2018prefrontal,igl2018deep}. We solve the POMDP by casting it as a Markov Decision Process (MDP) whose state space is the history of observation \(o_h\). We used a deep reinforcement learning method, called ACER, to find an approximately optimal policy for the POMDP \cite{wang2016sample}. 
For all reported experiments, we  built the environments within OpenAI Gym \cite{brockman2016openai} and used the OpenAI Baselines \footnote{https://github.com/openai/baselines} implementation of the deep RL algorithms.

%To discover the optimal strategy for this task problem, we use ACER \cite{wang2016sample}, which is a stable and sample efficient actor-critic deep reinforcement learning method. ACER learned to infer which option was most likely to have a maximum expected value while taking into account computational cost and noisy action outcomes.

\section{Results}
%\section{Results: Rational Analysis of Context effects}
% Context effect emerge from a computational rational process
% The results are split into four sections. In Section \ref{compare_benefitial}, Section \ref{prediction_data} and Section \ref{decision_process} we show that our new reinforcement learning model replicates the results of \cite{howes2016contextual} and then in Section \ref{number_actions}  we report new predictions that derive from the sequential nature of the reinforcement learner.

%The results are split into four sections. In Section \textbf{A}, \textbf{B} and \textbf{C} we show that our new reinforcement learning model replicates the results of \cite{howes2016contextual} and then in Section \textbf{D} we report new predictions that derive from the sequential nature of the reinforcement learner.

In order to test the model, we designed three different agents: 
The \textbf{integrated agent} could use both calculation and comparison selectively. States represent the results of calculation and comparison actions. The model can learn which observations are useful and not every observation needs to be made. There is no explicit integration of comparison and calculation. Instead, the results of comparison and calculation accumulate in the history and choice action values are conditional on these histories.
The \textbf{comparison-only agent} is same as the integrated agent but could only use comparison actions, and there are states only represent the comparison information.
The \textbf{calculation-only agent} is same as the integrated agent but could only use calculation actions and, there are states only represent the calculation information. The difference between the three models is the availability to use two kinds of observation information. All three agents learnt approximately optimal policies from experience. 

\begin{figure}[t]
\centering
\includegraphics[width=0.495\textwidth]{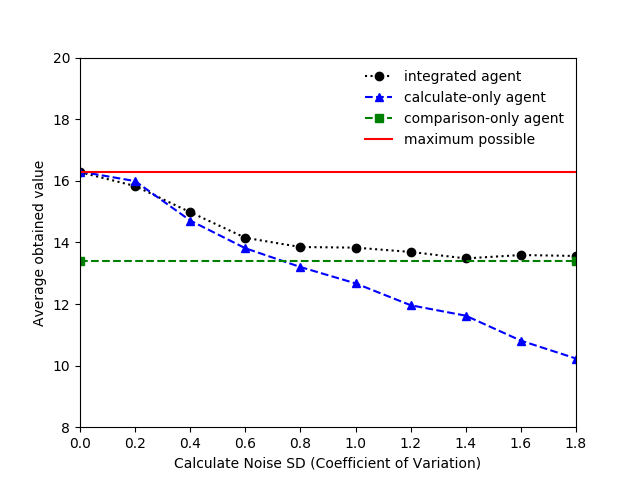} % Reduce the Figure size so that it is slightly narrower than the column.
\includegraphics[width=0.495\textwidth]{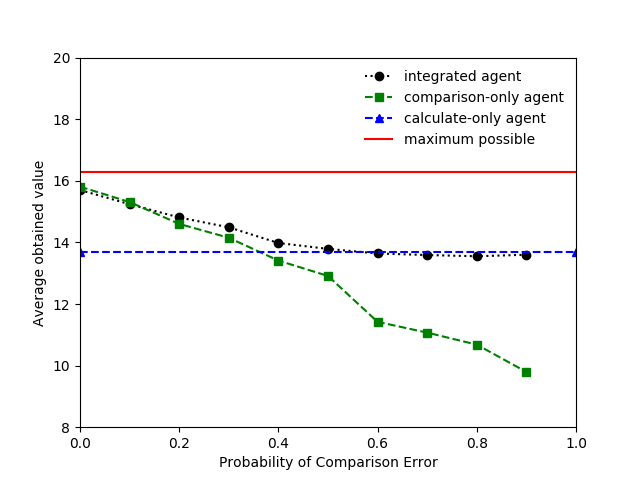}
\vspace{-2mm}
\caption{The mean expected value obtained by agents with different levels of noise: the coefficient of variation for the calculation noise (left panel) and the probability of comparison error (right panel). In the left panel, the comparison noise is fixed at \(P_{error}=0.3\). In the right panel, the calculation noise is fixed at \(\delta_{calc}=30\), that the coefficient of variation is 0.3. 
Results for 3 types of agent are presented in each panel: the comparison-only agent (green-doted line), calculation-only agent (blue-doted line) and integrated agent (black-doted line). This Figure replicates Figure 3 in \cite{howes2016contextual}.}
\label{maxU}
\end{figure}

% The difference between the three models is the availability to use two kinds of observation information. All three agents learnt approximately optimal policies from experience. 

In what follows we first show that our new reinforcement learning model replicates previous findings \cite{howes2016contextual} and then show that it makes new predictions derived from the sequential nature of the model.

\begin{figure*}[t]
\centering
\includegraphics[width=0.245\textwidth]{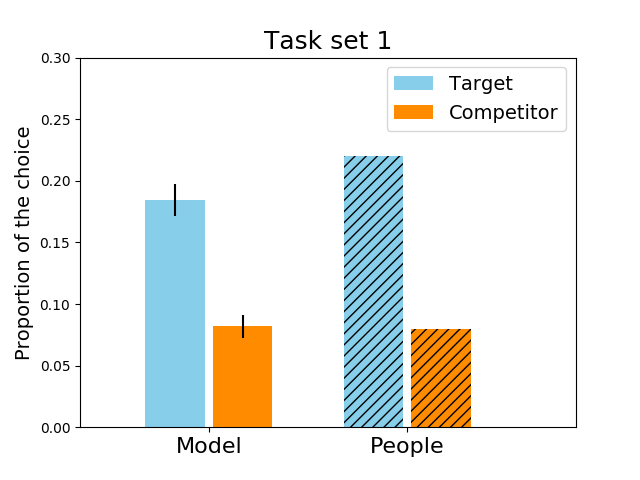} % Reduce the Figure size so that it is slightly narrower than the column.
\includegraphics[width=0.245\textwidth]{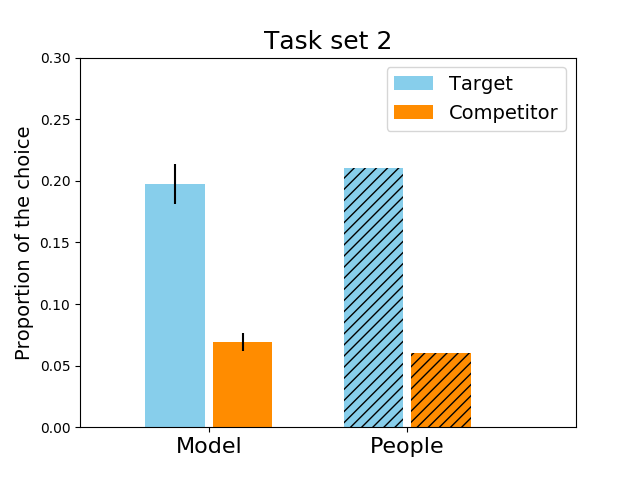}
\includegraphics[width=0.245\textwidth]{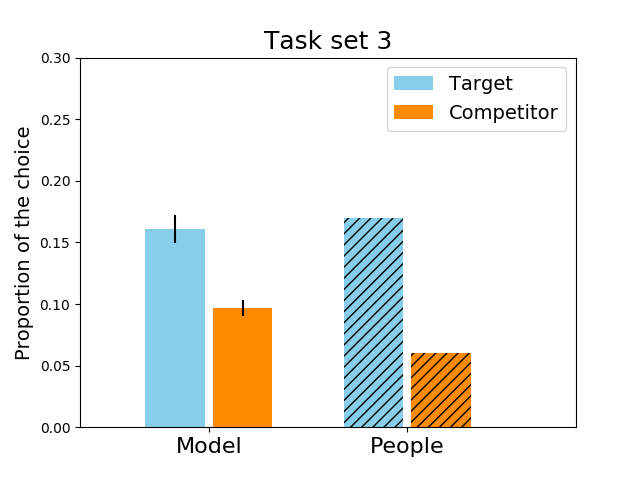}
\includegraphics[width=0.245\textwidth]{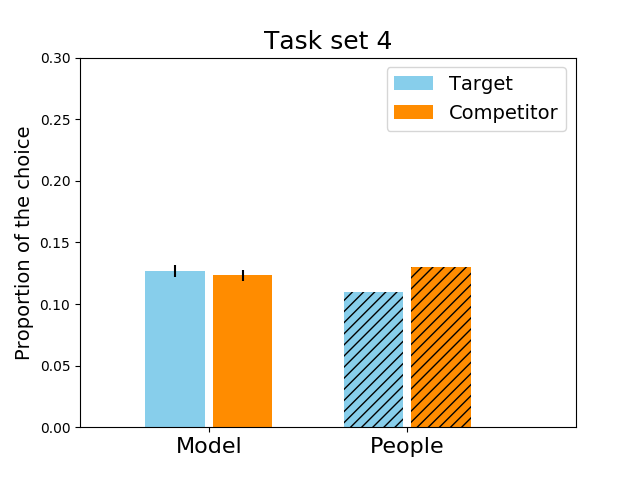}
\vspace{-5mm}
\caption{The integrated agent exhibits the attraction effect. A sample of agents was tested on each of four variants of the attraction effect task (in which the decoy is in slightly different positions). People and agent exhibit more target choices than Competitor choices in task sets 1, 2, and 3. As expected, neither the integrated agent, nor people, exhibit the effect in task set 4 where the decoy was not dominated by only one of the options and was therefore in a neutral position. Task 4 thereby acts as a control The human data is from \cite{wedell1991distinguishing}. The error bars indicate confidence interval (95\%) of the predictions made by the agent. This Figure replicates Figure 8 in \cite{howes2016contextual}.}
 \vspace{-5mm}
\label{wedell}
\end{figure*}

%We measured the utility of the different observations according to the expected reward of the resulting choice depend on such observation.

% ********* say how many trials of training. did it converge ***********

\textbf{Is it Beneficial to Compare Options?} 
In order to answer this question, we first fitted the distributions of the environment to those used in a prominent human experiment \cite{wedell1991distinguishing}. 
The probabilities \(p\) are \(\beta-\)distributed (\(a = 1, b = 1\)) and the values \(v\) are \(t-\)distributed (\(location=19.60, scale=5, degree \;of \; freedom=100 \)). For all the experiments below, we used the same distributions. Reported results are averaged over 10 runs, each with a different seed, after training on 3 million samples. All the details on setup and learning curves can be found in the supplementary material. 

All agents were tested with different levels of observation noise and the resulting performance is shown in Figure \ref{maxU}. 
The maximum expected value that could be achieved by any agent was 16.29 (horizontal upper bound in  Figure \ref{maxU}), which was calculated by averaging the maximum expected value of 3 options across 1 million choice sets sampled from the above distributions. 

In  Figure \ref{maxU} it can be seen that the integrated agent, using both calculation and comparison observations, can approximate the optimal policy when actions could be conducted without noise. Also, calculation-based and comparison-based agents are able to perform close to optimum when there is no noise. However, the noise has a negative effect on the performance of all types of agent. The average obtained value of choices decreases as noise increases. 

Figure \ref{maxU} also shows that the integrated agent combines the strengths of both noisy comparison and noisy calculation to make better decisions than the other agents in all noise conditions. The average expected value of the choices made by the integrated agent is greater than the other agents. In other words, the human-like integrated agent performs better in accumulating  reward than the agent that makes independent assessments of each option value. The results suggest that when there is observation uncertainty, both humans and artificial agents will gain higher reward if they compare options, rather than merely evaluate each option independently.

%This result also demonstrates that the comparison agents learnt an optimal policy in the choice task by trying to maximize expected value. Furthermore, the human-inspired model learns to make use of all available and valuable information to obtain higher expected rewards in the situation of uncertainty. It demonstrates that the resulting policy is a computational rational process.

% \subsection{Does the Integrated Agent Predict Human Performance?}
% \label{prediction_data}
\textbf{Does the Integrated Agent Predict Human Performance?} To determine whether the integrated agent (the agent that uses both comparison and calculation) predicts human performance, we measured its behaviour on the the attraction, compromise and similarity tasks. The human behaviour on these tasks is shown in Figure \ref{position}d. 
We used one fixed setting of the agent policy and parameter values. 
%Following this common rule, we use the proposed model to produce all the three context effects and one specific account of the attraction effects observed in human psychological experiments.  

\begin{figure}[t]
%\begin{wrapfigure}{r}{0.55\textwidth}
\centering
\includegraphics[width=1.0\columnwidth]{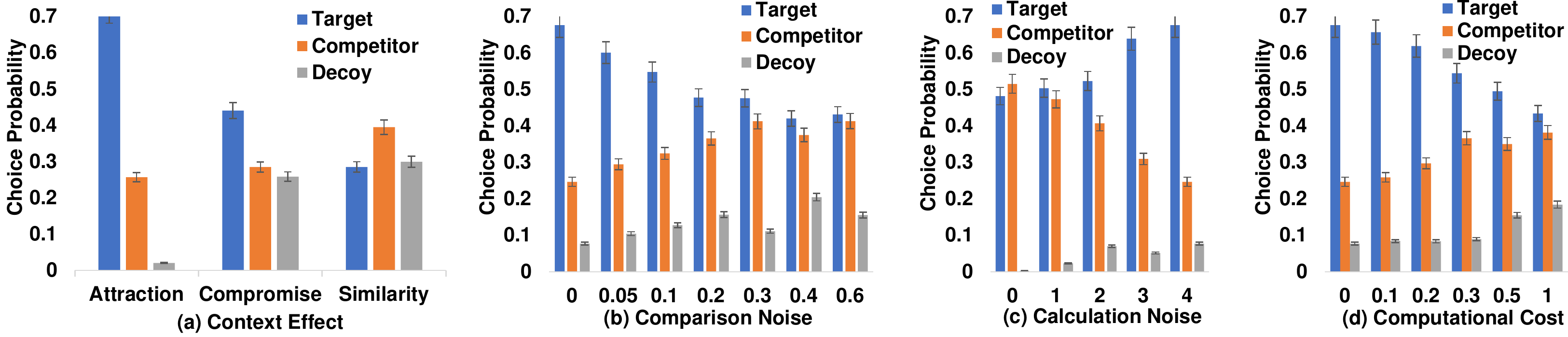}
\vspace{-2mm}
\caption{(a) The behaviour of the integrated agent for 3 types of context effect: attraction, compromise and similarity. (b)(c)(d) The effect of noise and computational cost on the contextual choice effect exhibited by the integrated agent. (b) Increased comparison noise reduces the effect size, (c) increased calculation noise increases the effect size, and (d) increased computational cost reduces the effect size. Error bars indicate  (95\%) confidence interval.}
\label{noise}
%\end{wrapfigure}
\end{figure} 

\begin{figure}[t]
\centering
\includegraphics[width=1.0\columnwidth]{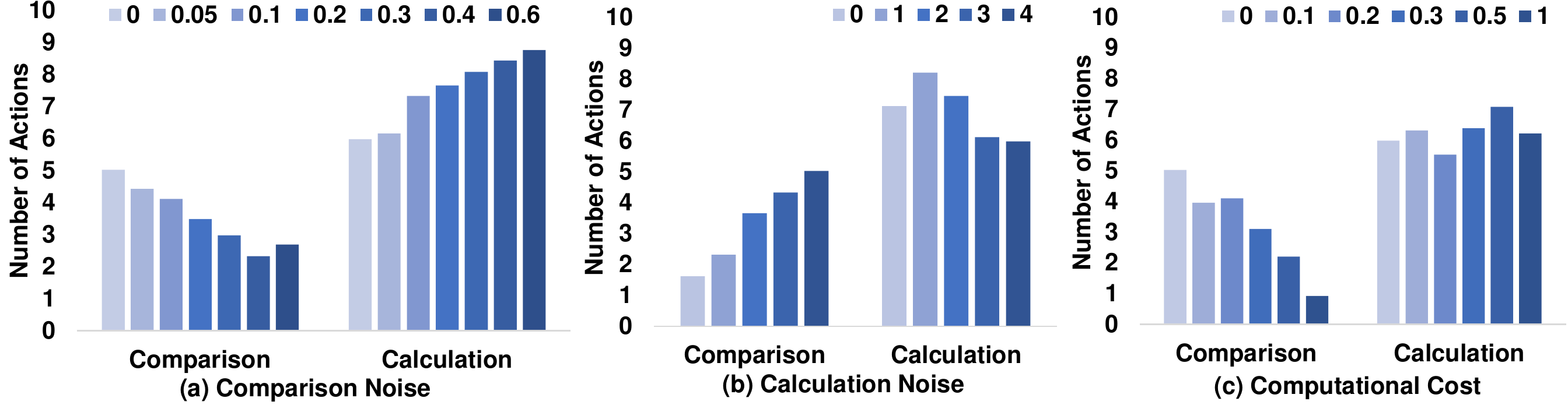}
\vspace{-2mm}
\caption{Effect of noise and computational cost on number of comparison and calculation actions taken. The analysis shows that the model chooses observations selectively depending on their utility.}
\label{action}
\end{figure} 

Agents were trained on  tasks which were randomly sampled from a \(\beta-\)distribution (\(a = 1, b = 1\)) for the probability \(p\) and \(t-\)distribution (\(location = 19.60, scale = 8.08, df = 100\)) for the value \(v\). After 3 million training samples, the agent converged and demonstrated stable performance. 
% We saved each agent at the end of training and tested them on two different choice sets. The average preference obtained by the model was the mean of the results by the saved models. 
The agent was repeatedly trained with adjusted values of the comparison noise, calculation noise, probability weighting parameters, cost of comparisons and calculation cost until the qualitative effects fitted the human performance (\cite{trueblood2012multialternative}; Figure \ref{position}d).
%\cite{busemeyer2019cognitive}. 
The fitted parameter values were: calculation noise \(\sigma_{calc}=4\), comparison error \(P(error_f)=0.1\), probability weighting parameters \(\alpha = 1\), the perceived cost of comparison \(C_{comparison} = -0.01\) and the calculation cost \(C_{calc}=-0.1\). We do not claim to have achieved the best possible fit, nor a better fit than other models. The point of the fit was to show that the qualitative effects exhibited by humans was within the space of behaviours generated by the agent.

%For fitting to theoretical predictions (\cite{busemeyer2019cognitive}; Figure 1 (B)), 
The results are averaged over 10 runs with different seed and shown in Figure \ref{noise}a. It shows that the agent generates the three context effects using one learnt policy and one fixed set of parameter values. Comparison of Figure \ref{noise}a to Figure \ref{position}d) shows that all of the qualitative effects are predicted. 

To further test the agent we fitted it to variations of the attraction effect in human performance
\cite{wedell1991distinguishing}. The fitted values were: calculation noise \(\sigma_{calc}=0.50\), comparison error \(P(error_f)=0.1\), probability weighting parameters \(\alpha = 1.5\), the perceive cost of comparison features \(C_{comparison} = -0.01\) and the calculation cost \(C_{calc}=-0.1\).
%
% TODO:
% 1. calculate the MSE or SSE.
% 2. analysis the information processing
% 3. plot the correlation between the accuracy and size of the effect
%
%
 %In total, 80 models were trained and tested. 
 %We tested the accuracy of each model using 10000 random sampled choice tasks. The results of 11 models were removed since their average accuracy was a very low 0.80, and were outliers from the others which had an average accuracy of 0.94. The 80 remaining models were tested on 50000*8*10 choice sets from the experiment. 
The results in Figure \ref{wedell} show that for both agents and people, the Target is selected more often than the Competitor in  three of the  task sets (1, 2, and 3). In contrast, and as expected, the Target and Competitor are selected equally often in the 4th task set by both agents and people. The decoy was positioned in a neutral position in task set 4 and does not therefore have an effect on the target choice rate.
 
 %As we have analysis in the last section, the solution to the POMDP is approximately optimal. It demonstrates that the context effect can emerge from approximately computationally rational processes. In the next section, we will investigate how the learning process adaptive to computational cost and cognitive limits.

% \subsection{Does the Uncertainty of Information Influence The Decision Process?}
% \label{decision_process}
\textbf{Does the Uncertainty of Information Influence the Decision Process?}
% \label{decision_process}
%Exiting cognitive process models vary from different assumption, which becomes a challenge of understanding the nature of context effect. They explain the effect in various way. We analysis the processing of context effect by testing the effect of potential influential factors. In this section, we will manipulate the different factor, computational cost and perceptual noise, which have effect on the size of preference reversals in the choice task.
We tested the consequences of noise on choice. The results  in Figure \ref{noise}b, c, d show that: (1) The size of attraction effect decreases as computational cost increases, (2) the attraction effect is weaker when the agent's accuracy of comparison is diminished with noise, (3) The effect is stronger when calculation noise is higher. 
While, there is no human data that directly tests the effect of  noise. There are a number of studies reporting that the rate of context effect diminishes with time pressure increases \cite{pettibone2012testing,trueblood2014multiattribute}.
As shown in Figure \ref{noise}b, c, d, the effects of time pressure on humans is consistent with the effect of increased  noise in the model. 

% \subsection{What is the Effect of Noise?}
% \label{number_actions}
\textbf{What are the Effects of Noise and Computation Cost?}
The effect of noise on the number of comparisons and calculation actions taken is shown in Figure \ref{action}. Increases in comparison noise leads to a selective reduction in the use of comparison and a selective increase in the use of calculation. Conversely, increases in calculation noise leads to a selective decrease in the use of calculation and an increase in the use of comparison. Increase in the cost of information gathering actions (comparison and expected value) reduces contextual effects on choice (Figure \ref{action}c) as less information is gathered.

%\emph{*** Are these consistent with human results? E.g. With Noguchi? What about Farmer?*** Are there time pressure effects? Look at papers on neural noise when people make choices}

%To analyze the model's predictions, we ran a 2-way ANOVA for context effect. The dependent variable is the proportion of the choice that the agent chose 

%The size of context effect decrease as increasing comparison noise or computation cost. It is consistent to human data that with decrease of deliberate time.

%\subsection{Does the model predict selective attention?}

%The majority trajectories of the process show that the agent gathers information about options selectively rather than sampling randomly. This is consistent to 'less is more' heuristic of human cognition. And also correspond to empirical evidence that people use selective attention in decision making.

\section{Discussion}

%In the work, we investigate the core mechanisms giving rise to context effect.  

%The economic decision making is about how to choose options based on reinforcement history (Building bridges between perceptual and economic decision making: neural and computational mechanisms)

%How to account for context sensitivity of valuation (Where Does Value Come From?) So evaluate the information of sensors based on relative value, which combing economic, cognitive and neural basis, rather than on absolutely value. naïve utility calculus

%We have replicated the findings of \cite{howes2016contextual} by demonstrating that an agent using the comparison actions that underpinning apparent human biases generates higher  reward than an agent that makes independent assessments of value. 
While ours is not the first work to demonstrate the rationality of preference reversal phenomena \cite{howes2016contextual}, nor the first work to use POMDPs to model humans \cite{daw2006representation,rao2010decision}), 
it is the first to formulate the contextual choice problem as a POMDP and demonstrate that a reinforcement learning agent that uses \emph{comparison} observations generates higher reward than an agent that makes independent assessments of value. These comparison actions, when deployed by people, have been thought by many to violate independence axioms. They have been shown to underpin preference reversals in humans \cite{noguchi2014attraction}. As has previously been pointed out, this seemingly paradoxical result makes sense when it is appreciated that the comparison of options reduces the uncertainty of option values. 

Further, extending \cite{howes2016contextual}, we have demonstrated that  the same pattern of behaviours that are thought to be irrational in humans, will emerge from a \emph{learning} process that  attempts to maximize the cumulative reward of action. 
%The reinforcement learner solves simple three option sequential decision tasks formulated as a POMDP. 
Further, our results show that comparison actions are \emph{preferred} by the agent as observation noise increases (in previous models comparisons have been assumed) and we have also shown that higher information gathering costs can diminish the use of comparisons and reduce the preference reversal rate; thereby extending previous analysis to account for the economics of information gathering in contextual choice tasks.

The approach that we have taken in this paper is an example of a broader class of analysis known as Computational Rationality \cite{lewis2014computational,gershman2015computational,lieder2019resource}. This approach starts from the assumption that people are approximately rational given the bounds imposed by the computation required for cognition \cite{lewis2014computational,lieder2017automatic,howes2009rational}. It then seeks to discover the computational limits that give rise to boundedly optimal \cite{russell1994provably} but apparently irrational behaviours. This aim demands that the analyst derive bounded optimal policies for well-formed decision problems. 
%This means finding bounded optimal trade-offs between expected utility and the cost of gathering information. 
Our results suggest an answer to the paradox of why it is worth motivating  machine learning algorithms with apparently biased human decision making. While the behaviour appears biased, the underlying processes and heuristics (e.g. the use of option comparison) lead to gains in efficiency and therefore reward.

%What is more, our results contribute to a growing body of work \cite{gigerenzer2018bias,lieder2017strategy,chen2015optimal} calling into question the long list of apparent irrationalities reported in the Economic literature. More may be amenable to POMDP, Meta-MDP, or MDP explanations and turn out to be rational adaptations to environmental and cognitive limits. 
%In addition, there is important work on heuristics e.g. the recognition heuristic, which can be explained with MDPs and are likely to be helpful for ML (Lieder. An automatic method for discovering rational heuristics for risky choice.). 

\section{Conclusion}

Machine learning researchers can take inspiration from apparent human irrationalities. This claim is supported by our demonstration that reinforcement learning agents that seek to maximize cumulative reward when observations are uncertain can improve performance by selectively comparing option values and not merely making independent assessments of each option. This human-like processing appears irrational but is demonstrably rational under bounds imposed by uncertainty in the observation function. 

\newpage
\clearpage

\section*{Broader Impact}
The current work is potentially influential on the future of how RL is used in systems designed to understand and interact with humans. It may have positive ethical implications by virtue of the fact that it enhances the scientific understanding of the relationship between human and machine information gathering and decision processes. 
%The idea that people are irrational has led to a programme of research into how to ``debias" human decision strategies. But, if it is correct that decision strategies are adapted rationally to cognitive and environmental constraints, then efforts to debias decision strategies might be harmful and lead to worse decisions. Our demonstration that, by learning an optimal policy, a reinforcement learning agent can produce the same apparent irrationalities as people suggests deeper thought is needed to understand the causes of apparent bias. 
We have no reason to believe that the data used to train and validate the model is biased in a way that would disadvantage  protected or minority groups on the basis of race or gender.

%Authors are required to include a statement of the broader impact of their work, including its ethical aspects and future societal consequences. 
%Authors should discuss both positive and negative outcomes, if any. For instance, authors should discuss a) 
%who may benefit from this research, b) who may be put at disadvantage from this research, c) what are the consequences of failure of the system, and d) whether the task/method leveragesbiases in the data. If authors believe this is not applicable to them, authors can simply state this.

\medskip

\small

\bibliographystyle{unsrt} % Use for unsorted references  
\bibliography{reference}

\end{document}